\documentclass[11pt,a4paper]{article}
\usepackage[hyperref]{eacl2021}
\usepackage{times}
\usepackage{latexsym}

\usepackage{microtype}

\usepackage{amsmath}
\usepackage{booktabs}

\usepackage{graphicx}

\aclfinalcopy 
\def\aclpaperid{198} 

\title{Does Typological Blinding Impede Cross-Lingual Sharing?}

\author{Johannes Bjerva \\
  Department of Computer Science \\
  Aalborg University \\
  \texttt{jbjerva@cs.aau.dk} \\\And
  Isabelle Augenstein \\
  Department of Computer Science \\
  University of Copenhagen \\
  \texttt{augenstein@di.ku.dk} \\}

\date{}

\begin{document}
\maketitle
\begin{abstract}
Bridging the performance gap between high- and low-resource languages has been the focus of much previous work.
Typological features from databases such as the World Atlas of Language Structures (WALS) are a prime candidate for this, as such data exists even for very low-resource languages.
However, previous work has only found minor benefits from using typological information.
Our hypothesis is that a model trained in a cross-lingual setting will pick up on typological cues from the input data, thus overshadowing the utility of explicitly using such features.
We verify this hypothesis by blinding a model to typological information, and investigate how cross-lingual sharing and performance is impacted.
Our model is based on a cross-lingual architecture in which the latent weights governing the sharing between languages is learnt during training. 
We show that (i) preventing this model from exploiting typology severely reduces performance, while a control experiment reaffirms that (ii) encouraging sharing according to typology somewhat improves performance.
\end{abstract}

\section{Introduction}

Most languages in the world have little access to NLP technology due to data scarcity \citep{joshi2020state}.
Nonetheless, high-quality multilingual representations can be obtained using only a raw text signal, e.g.~via multilingual language modelling \citep{bert}.
Furthermore, structural similarities of languages are to a large extent documented in typological databases such as the World Atlas of Language Structures (WALS, \citet{wals}).
Hence, developing models which can take use typological similarities of languages is an important direction in order to alleviate language technology inequalities.

\begin{figure}[tb]
    \centering
    \includegraphics[width=1.0\columnwidth]{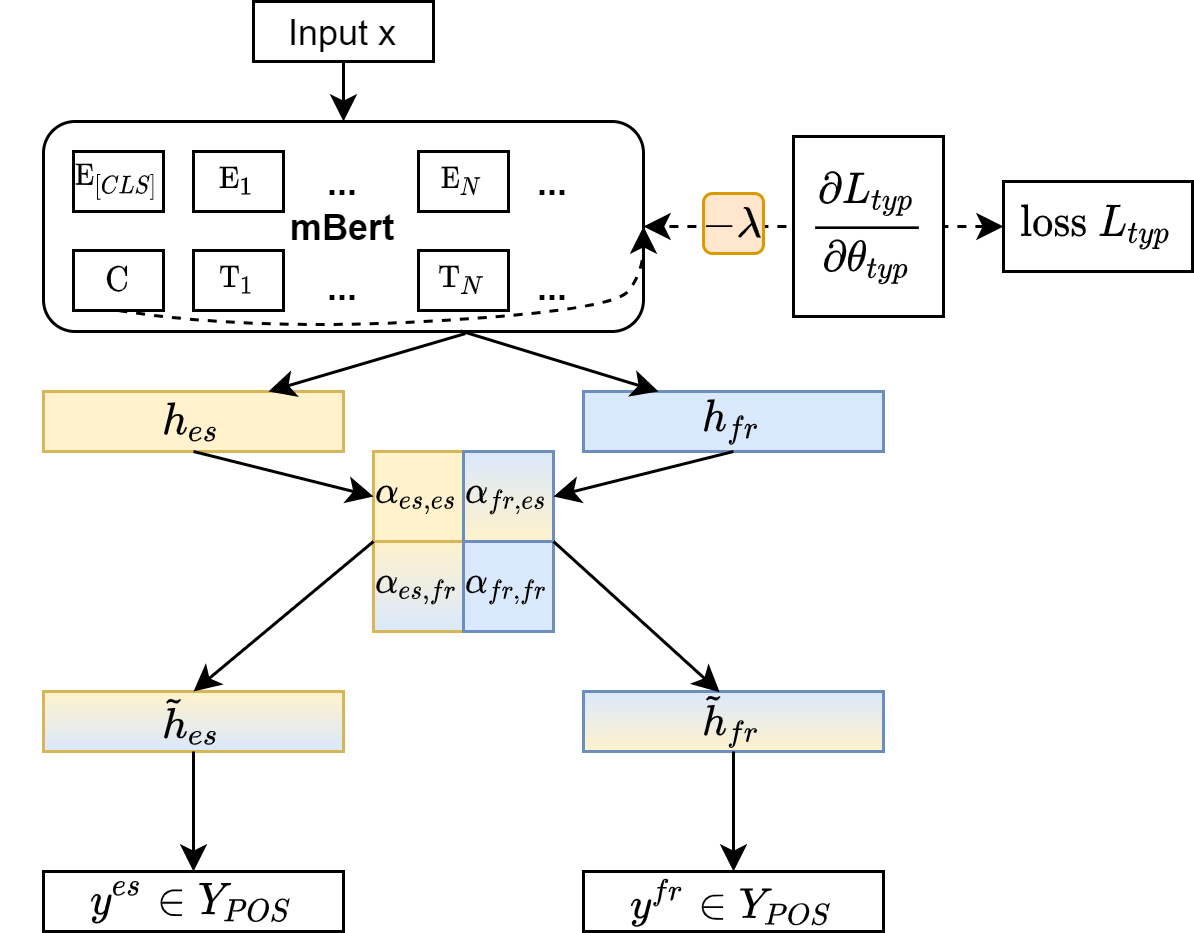}
    \caption{A PoS tagger is exposed (or blinded with gradient reversal, $-\lambda$) to typological features. 
    Observing $\alpha$ values tells us how typology affects sharing.}
    \label{fig:model}
\end{figure}

While previous work has attempted to use typological information to inform NLP models, our work differs significantly from such efforts in that we \textit{blind} a model to this information.
Most previous work includes language information as features, by using language IDs, or language embeddings (e.g.~\citet{ammar2016many,o2016survey,ostling-tiedemann-2017-continuous,ponti2019modeling,oncevay2020bridging}).
Notably, limited effects are usually observed from including typological features explicitly.
For instance, \citet{delhoneux2018parameter} observe positive cross-lingual sharing effects only in a handful of their settings.
We therefore hypothesise that relevant typological information is learned as a by-product of cross-lingual training. 
Hence, although models do benefit from this information, it is not necessary to provide it explicitly in a high-resource scenario, where there is abundant training data.
This is confirmed by \citet{bjerva2018phonology}, who find that, e.g., language embeddings trained on a morphological task can encode morphological features from WALS.

In contrast with previous work, we \textit{blind} a model to typological information, by using adversarial techniques based on gradient reversal \citep{ganin2014unsupervised}.
We evaluate on the structured prediction and classification tasks in XTREME \citep{xtreme}, yielding a total of 40 languages and 4 tasks.
We show that when a model is blinded to typological signals relating to syntax and morphology, performance on related NLP tasks drops significantly.
For instance, the mean accuracy across 40 languages for POS tagging drops by 1.8\% when blinding the model to morphological features.


\section{Model}
An overview of the model is shown in Figure~\ref{fig:model}. We model each task in this paper using the following steps. 
First, contextual representations are extracted using multilingual BERT (m-BERT, \citet{bert}), a transformer-based model \citep{vaswani2017attention}, trained with shared word-pieces across languages.
We either blind m-BERT to typological features, with an added adversarial component based on gradient reversal \citep{ganin2014unsupervised}, or expose it to them via multi-task learning (MTL, \citep{caruana:1997}).
Representations from m-BERT are fed to a latent multi-task architecture learning network \citep{ruder2019latent}, which includes $\alpha$ parameters we seek to investigate.
The model learns which parameters to share between languages (e.g.~$\alpha_{es,fr}$ denotes sharing between Spanish and French).

\subsection{Sharing architecture}

Our sharing architecture is based on that of \citet{ruder2019latent}, which has latent variables learned during training, governing which layers and subspaces are shared between tasks, to what extent, as well as the relative weighting of different task losses.
We are most interested in the parameters which control the sharing between the hidden layers allocated to each task, referred to as $\alpha$ parameters \citep{ruder2019latent}.
Consider a setting with two tasks $A$ and $B$. 
The outputs $h_{A, k}$ and $h_{B, k}$ of the $k$-th layer for task $A$ and $B$ interact through the $\alpha$ parameters, for which the output is defined as: 

\begin{equation}
\begin{bmatrix}
\widetilde{h}_{A, k}\\
\widetilde{h}_{B, k}
\end{bmatrix} = 
\begin{bmatrix}
           \alpha_{AA} & \alpha_{AB} \\
           \alpha_{BA} & \alpha_{BB}
         \end{bmatrix} 
\begin{bmatrix}
{h_{A, k}}^\top \: , & {h_{B, k}}^\top
\end{bmatrix}
\label{eq:alphas}
\end{equation}

where $\widetilde{h}_{A, k}$ is a linear combination of the activations for task $A$ at layer $k$, weighted with the learned $\alpha$s.
While their model is an MTL model, we choose to interpret this differently by considering each \textit{language} as a task, yielding $\alpha\in \mathbf{R}^{l\times l}$, where $l$ is the number of languages for the given task.
Each activation $\widetilde{h}_{A, k}$ is then a linear combination of the language specific activations $h_{A, k}$.
These are used for prediction in the downstream tasks, as in the baselines from \citet{xtreme}.

Crucially, this model allows us to draw conclusions about parameter sharing between languages by observing the $\alpha$ parameters under the blinding and prediction conditions.
We will combine this insight with observing downstream task performance in order to draw conclusions about the effects of typological feature blinding and prediction.


\subsection{Blinding/Exposing a Model to Typology}

We introduce a component which can either blind or expose the model to typological features.
We implement this as a single task-specific layer per feature, using the \texttt{[CLS]} token from m-BERT model, without access to any of the soft sharing between languages from $\alpha$-layers.
Each layer optimises a categorical cross-entropy loss function ($L_{typ}$).

For this task, we predict typological features drawn from WALS \citep{wals}, inspired by previous work \citep{bjerva2018phonology}.
Unlike previous work, we also blind the model to such features by including a gradient reversal layer \citep{ganin2014unsupervised}, which multiplies the gradient of the typological prediction task with a negative constant ($-\lambda$), inspired by previous work on adversarial learning \citep{goodfellow2014explaining,zhang2019limitations, chen2019stateful}.
We hypothesise that using a gradient reversal layer for typology will yield typology-invariant features, and that this will perform worse on tasks for which the typological feature at hand is important.
For instance, we expect that blinding a model to syntactic features will severely reduce performance for tasks which rely heavily on syntax, such as POS tagging.

\section{Cross-Lingual Experiments}
We investigate the effects of typological blinding, using typological parameters as presented in WALS \citep{wals}.
The experiments are run on XTREME \citep{xtreme}, which includes up to 40 languages from 12 language families and two isolates. We experiment on the following languages (ISO 639-1 codes): af, ar, bg, bn, de, el, en, es, et, eu, fa, fi, fr, he, hi, hu, id, it, ja, jv, ka, kk, ko, ml, mr, ms, my, nl, pt, ru, sw, ta, te, th, tl, tr, ur, vi, yo, and zh.
We experiment on four tasks: POS (part of speech tagging), NER (named entity recognition), XNLI (cross-lingual natural language inference), and PAWS-X (paraphrase identification).
Our general setup for the structured prediction tasks (POS and NER) is that we train on all available languages, and downsample to 1,000 samples per language.
For the classification tasks XNLI and PAWS-X, we train on the English training data and fine-tune on the development sets, as no training data is available for other languages.
Hence, typological differences will be the main factor in our results, rather than differences in dataset sizes.

\subsection{Typological Prediction and Blinding}
We first investigate whether prohibiting or allowing access to typological features has an effect on model performance using our architecture.
We hypothesise that our multilingual model will leverage signals related to the linguistic nature of a task when optimising its its sharing parameters $\alpha$.

There exists a growing body of work on prediction of typological features \citep{daume:2007,murawaki:2017,bjerva_augenstein:18:iwclul,bjerva-etal-2019-probabilistic,bjerva-etal-2019-uncovering}, most notably in a recent shared task on the subject \citep{bjerva-etal-2020-sigtyp}. 
While we are inspired by this direction of research, our contribution is not concerned with the accuracy of the prediction of such features, and this is therefore not evaluated in detail in the paper.

Moreover, an increasing amount of work measures the correlation of predictive performance of cross-lingual models with typological features as a way of probing what a model has learned about typology \cite{malaviya-etal-2017-learning,journals/corr/abs-2009-12862,gerz-etal-2018-relation,nooralahzadeh2020meta,zhao2020inducing}. In contrast to such post-hoc approaches, our experimental setting allows for measuring the impact of typology on cross-lingual sharing performance in a direct manner as part of the model architecture.

\paragraph{Syntactic Features}
We first blind/expose the model to syntactic features 
 from WALS \citep{wals}.
We take the set of word order features which are annotated for all languages in our experiments, resulting in 33 features.
This includes features such as \textit{81A: Order of Subject, Object and Verb}, which encodes what the preferred word ordering is (if any) in a transitive clause.
For all features, we exclude feature values which do not occur for our set of languages.
We hypothesise that performance will drop for all four tasks, as they all require syntactic understanding.

\paragraph{Morphological Features}
We next attempt to blind/expose the model to the morphological features in WALS.
We use the same approach as above, resulting in a total of 8 morphological features.
This includes features such as \textit{26A: Prefixing vs. Suffixing in Inflectional Morphology}, indicating to what extent a language uses prefixing or suffixing morphology.
We hypothesise that mainly the POS tagging task will suffer under this condition, whereas other tasks only to some extent require morphology.

\paragraph{Phonological Features} 
We next consider a control experiment, in which we attempt to blind/expose the model to phonological features in WALS.
We arrive at a total of 15 phonological features, such as \textit{1A: Consonant Inventories} which indicates the size of the consonant inventory of a language.
We expect the performance to remain relatively unaffected by this task, as phonology ought to have little importance given a textual input.

\paragraph{Genealogical Features}
Finally, we attempt to use what one might consider to be language meta-data.
We attempt to blind/expose the model to what language family a language belongs to.
This can be seen as a type of proxy to language similarity, and correlates relatively strongly with structural similarities in languages. 
Because of this correlation with structural similarities, we expect blinding under this condition to only slightly reduce performance for all tasks, as previous work has shown this type of relationship not to be central in language representations \citep{bjerva:cl:2019}.

\subsection{Results}
In general, we observe a drop in performance when blinding the model to relevant typological information, and an increase in performance when exposing the model to it (Table~\ref{tab:prediction}).
For phonological blinding or prediction, none of the four tasks is noticeably affected.
Although, e.g., both the syntactic and morphological prediction tasks increase performance on POS tagging, it is not straightforward to draw conclusions on which of these is the most efficient, as there is a substantial correlation between syntactic and morphological features.
As for XNLI and PAWS-X, performance notably drops under both the syntactic and genealogical blinding tasks.

\begin{figure*}[tbh]
    \centering
    \includegraphics[width=\textwidth]{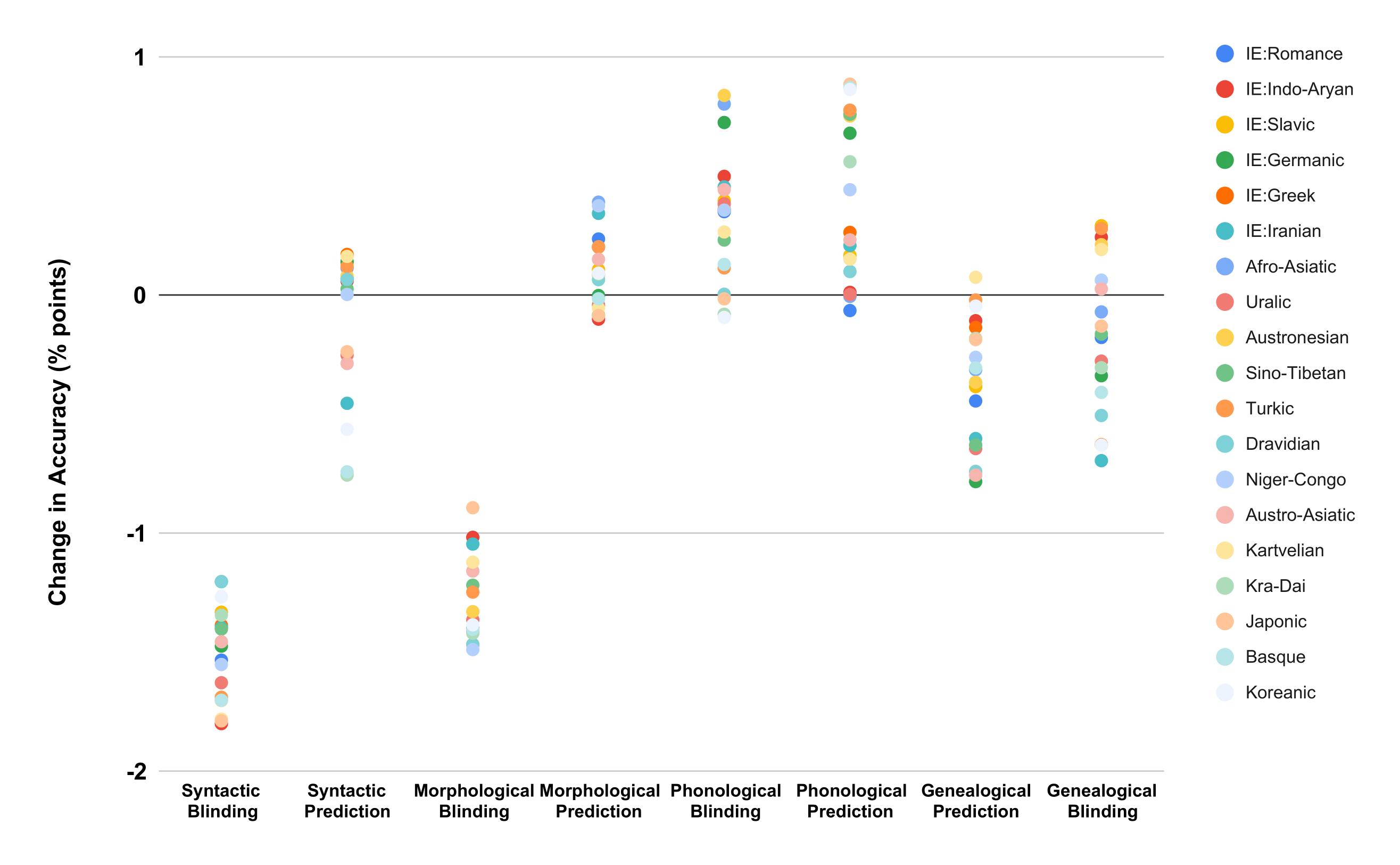}
    \caption{PoS tagging results per language family across blinding and prediction conditions}
    \label{fig:lang_families}
\end{figure*}

Figure~\ref{fig:lang_families}  shows results for PoS tagging under prediction and blinding across language families, following the same scheme as 
\citet{xtreme}.
Interestingly, 
the syntactic and morphological blinding settings are robust across all language families, yielding a drop in accuracy across the board.
All other conditions yield mixed results.
This further strengthens our argument that preventing a model from learning syntactic and morphological features can be severely detrimental.

\setlength{\tabcolsep}{4pt}
\begin{table}[t]
    \centering
    \resizebox{1.0\columnwidth}{!}{
    \begin{tabular}{lllll}
        \toprule
        \textbf{Model}          & \textbf{POS} & \textbf{NER} & \textbf{XNLI} & \textbf{PAWS-X}  \\
        \midrule
        + Syntactic Blind.      & 85.3$^{-}$ & 76.4 & 64.2$^{-}$ & 80.6$^{-}$ \\
        + Morphological Blind.  & 85.0$^{-}$ & 77.2 & 64.9 & 81.4 \\
        + Phonological  Blind.  & 86.7 & 77.1 & 65.0 & 81.6 \\
        + Genealogical  Blind.  & 86.1 & 77.0 & 64.7 & 81.1 \\
        \midrule
        \textit{m-BERT  baseline}        & 86.8 & 77.3 & 65.1 & 81.7 \\
        \midrule
        + Syntactic Pred.       & 87.0 & 77.5 & \textbf{65.3$^{+}$} & \textbf{81.9$^{+}$} \\
        + Morphological Pred.   & \textbf{87.2$^{+}$} & 77.3 & 65.2 & 81.7 \\
        + Phonological  Pred.   & 86.7 & 77.1 & 65.0 & 81.7 \\
        + Genealogical  Pred.   & 87.0 & \textbf{77.6} & \textbf{65.3$^{+}$} & 81.8 \\
        \bottomrule
    \end{tabular}
    }
    \caption{Typological Blinding and Prediction. Mean POS accuracy, NER F1 scores, XNLI accuracy and PAWS-X accuracy across all languages. $^{+}$ and $^{-}$ indicate significantly better or worse performance respectively, as determined by a one-tailed t-test ($p<0.01$).}
    \label{tab:prediction}
\end{table}
\setlength{\tabcolsep}{6pt}

\subsection{The Effect of Typology on Latent Architecture Learning}

The results show that preventing access to typological features hampers performance, whereas providing access improves performance.
We now turn to an analysis of how the model shares parameters across languages in this setting. 
Our hypothesis is that blinding will prevent models from sharing parameters between similar languages, in spite of typological similarities.
Concretely, we expect that the drop in POS tagging performance under morphological blinding is caused by lower $\alpha$ weights between languages which are morphologically similar, and higher $\alpha$ weights between languages which are dissimilar.
Recall that these parameters are latent variables learned by the model, regulating the amount of sharing between languages (see Eq.~\ref{eq:alphas}).
We investigate the correlations between the $\alpha$ sharing parameters, and two proxies of language similarity. 
We focus on the POS task, as the results from the typological blinding and prediction experiments were the most pronounced here, as both morphological and syntactic blinding affected performance.

Our first measure of language similarity is based on \citet{bjerva:cl:2019}, who introduce what they refer to as structural similarity.
This is based on dependency statistics from the Universal Dependencies treebank \citep{ud26}, resulting in vectors which describe how different syntactic relations are used in each language.
Previous work has shown that this measure of similarity correlates strongly with that learned in embedded language spaces during multilingual training.
In addition to considering these dependency statistics, we also use language embeddings drawn form \citet{ostling-tiedemann-2017-continuous}.
For each language similarity measure we calculate its pairwise Pearson correlation with the $\alpha$ values learned under each condition.

Table~\ref{tab:sluice_analysis} shows correlations between $\alpha$ weights and similarities increase when predicting typological features, and decreases when blinded to such features.
Hence, when the model has indirect access to, e.g., the SVO word ordering features of languages, sharing also reflects this.

\begin{table}[t]
    \centering
    \resizebox{1.0\columnwidth}{!}{
    \begin{tabular}{lrr}
        \toprule
        \textbf{Model}            & \textbf{Struct.} & \textbf{Lang. Emb.} \\
        \midrule
        Syntactic Blind.        & 0.31 & 0.27 \\
        Morphological Blind.    & 0.34 & 0.29 \\
        Phonological  Blind.    & 0.40 & 0.41 \\
        Genealogical  Blind.    & 0.29 & 0.31 \\
        \midrule
        No blind./pred.             & 0.43    & 0.40    \\
        \midrule
        Syntactic Pred.         & \textbf{0.52} & 0.53 \\
        Morphological Pred.     & 0.49 & \textbf{0.56} \\
        Phonological  Pred.     & 0.41 & 0.39 \\
        Genealogical  Pred.     & 0.47 & 0.38 \\
        \bottomrule
    \end{tabular}
    }
    \caption{Pearson correlations between $\alpha$ weights and language similarity measures.}
    \label{tab:sluice_analysis}
\end{table}

\section{Discussion}
We have shown that blinding a multilingual model to typological features severely affects sharing across a relatively large language sample, and for several NLP tasks.
The effects on model performance, as evaluated over 40 languages and 4 tasks from XTREME \citep{xtreme}, were the largest for POS tagging.
The fact that smaller effects were observed for NER, could be because this task relies more on memorising NEs rather than using (morpho-)syntactic cues \citep{augenstein2017}.
Furthermore, the relatively small effects on XNLI and PAWS-X can also be interpreted as evidence for that typology is less important in these tasks than in more traditional linguistic analysis.

A potential critique of our approach is that it merely blinds the model to language identities.
This could be the case, if only some latent representation of, e.g.,  ``SVO'' ordering is used to represent a language identity. 
However, previous work has shown that morphological information is encoded by the type of model we investigate.
Hence, since we only blind features in a single category at a time, we expect that the model's representation of language identities is unaffected. 

Not only do we observe a drop in performance when blinding a model to syntactic features, but we also observe that the $\alpha$ sharing weights in our model do not appear to correlate with linguistic similarities in this setting.
Conversely, encouraging a model to consider typology, by jointly optimising it for typological feature prediction, improves performance in general.
Furthermore, $\alpha$ weights in this scenario converge towards correlating with structural similarities of languages.
This is in line with recent work which has found that m-BERT uses fine-grained syntactic distinctions in its cross-lingual representation space \citep{chi:2020}.

We interpret this as evidence for the fact that typology can be a necessity for modelling in NLP.
Our results furthermore corroborate previous work in that we only find moderate benefits from including typological information explicitly.
We expect that this to a large degree is due to the typological similarities of languages being encoded implicitly based on correlations between patterns in the input data.
As low-resource languages often do not even have access to any substantial amount of raw text, but often do have annotations in WALS, we expect that using typological information can go some way towards building truly language-universal models.

\section{Conclusions}
We have shown that preventing access to typology can impede the performance of cross-lingual sharing models.
Investigating latent weights governing the sharing between languages shows that this prevents the model from sharing between typologically similar languages, which is otherwise learned based on patterns in the input.
We therefore expect that using typological information can be of particular interest for building truly language-universal models for low-resource languages.

\section*{Acknowledgements}
This research has received funding from the Swedish Research Council (grant No 2019-04129), and  the NVIDIA Corporation (Titan Xp GPU).

\clearpage

\bibliographystyle{acl_natbib}
\bibliography{anthology,emnlp2020}




\end{document}